\documentclass[sigconf]{acmart}
\copyrightyear{2025}
\acmYear{2025}
\setcopyright{cc}
\setcctype{by}
\acmConference[SIGIR '25]{Proceedings of the 48th International ACM SIGIR Conference on Research and Development in Information Retrieval}{July 13--18, 2025}{Padua, Italy}
\acmBooktitle{Proceedings of the 48th International ACM SIGIR Conference on Research and Development in Information Retrieval (SIGIR '25), July 13--18, 2025, Padua, Italy}
\acmDOI{10.1145/3726302.3730007}
\acmISBN{979-8-4007-1592-1/2025/07}

\usepackage{balance} 
\usepackage{subfigure}
\usepackage{hyperref}
\usepackage[capitalise]{cleveref}
\usepackage{adjustbox}
\usepackage{algpseudocode}
\floatname{algorithm}{Procedure}

\usepackage{algorithm}
\usepackage{makecell}
\usepackage{colortbl}
\usepackage{array} 
\usepackage{xcolor}  
\usepackage{multicol}
\usepackage{multirow}
\usepackage{ulem}    

\usepackage[skip=3.5pt]{caption}

\newcommand{\corrauthor}{Bingbing Xu}

\AtBeginDocument{%
 }

\begin{document}



\title{InfoNCE is a Free Lunch for \\ Semantically guided Graph Contrastive Learning}
\author{Zixu Wang}
\orcid{0009-0006-1327-6366}
\affiliation{%
  \institution{State Key Laboratory of AI Safety, Institute of Computing Technology, Chinese Academy of Sciences}
  \department{University of Chinese Academy of Sciences}
  \city{Beijing}
  \country{China}
}
\email{wangzixu22s@ict.ac.cn}

\author[cor]{Bingbing Xu}
\orcid{0000-0002-0147-2590}
\affiliation{%
  \institution{State Key Laboratory of AI Safety, Institute of Computing Technology, Chinese Academy of Sciences}
  \city{Beijing}
  \country{China}
}
\email{xubingbing@ict.ac.cn}
\authornote{Corresponding author: \corrauthor}

\author{Yige Yuan}
\orcid{0000-0001-8856-668X}
\affiliation{%
  \institution{State Key Laboratory of AI Safety, Institute of Computing Technology, Chinese Academy of Sciences}
  \department{University of Chinese Academy of Sciences}
  \city{Beijing}
  \country{China}
}
\email{yuanyige20z@ict.ac.cn}

\author{Huawei Shen}
\orcid{0000-0003-2425-1499}
\affiliation{%
  \institution{State Key Laboratory of AI Safety, Institute of Computing Technology, Chinese Academy of Sciences}
  \city{Beijing}
  \country{China}
}
\email{shenhuawei@ict.ac.cn}

\author{Xueqi Cheng}
\orcid{0000-0002-5201-8195}
\affiliation{%
 \institution{State Key Laboratory of AI Safety, Institute of Computing Technology, Chinese Academy of Sciences}
  \city{Beijing}
  \country{China}
}
\email{cxq@ict.ac.cn}



\begin{abstract}

As an important graph pre-training method, Graph Contrastive Learning (GCL) continues to play a crucial role in the ongoing surge of research on graph foundation models or LLM as enhancer for graphs.
Traditional GCL optimizes InfoNCE by using augmentations to define self-supervised tasks, treating augmented pairs as positive samples and others as negative. However, this leads to semantically similar pairs being classified as negative, causing significant sampling bias and limiting performance.
In this paper, we argue that GCL is essentially a Positive-Unlabeled (PU) learning problem, where the definition of self-supervised tasks should be semantically guided, i.e., augmented samples with similar semantics are considered positive, while others, with unknown semantics, are treated as unlabeled. From this perspective, the key lies in how to extract semantic information.
To achieve this, we propose IFL-GCL, using InfoNCE as a "free lunch" to extract semantic information. 
Specifically, We first prove that under InfoNCE, the representation similarity of node pairs aligns with the probability that the corresponding contrastive sample is positive. 
Then we redefine the maximum likelihood objective based on the corrected samples, leading to a new InfoNCE loss function.
Extensive experiments on both the graph pretraining framework and LLM as an enhancer show significantly improvements of IFL-GCL in both IID and OOD scenarios, achieving up to a 9.05\% improvement, validating the effectiveness of semantically guided.
Code for IFL-GCL is publicly available at: https://github.com/Camel-Prince/IFL-GCL.


\end{abstract}





\maketitle

\section{Introduction}\label{intro}
As a flexible and powerful data structure, graphs play a critical role in modeling domains such as social networks\cite{liu2021content,bao2013cumulative, shen2011exploring}, chemical molecules\cite{wang2023graph}, and transportation systems\cite{rahmani2023graph}. 
As a prominent paradigm for pre-training graph models, Graph Contrastive Learning(GCL) \cite{BGRL,costa, GCC, GRACE, MVGRL, GCA} learns representations that effectively capture both feature and structural information in a self-supervised manner, which continues to play an indispensable role in the wave of graph foundation model.
For example, graph contrastive learning is a commonly used self-supervised pre-training method in the paradigm where LLM serves as an enhancer and graph model as the core\cite{gaugllm, llms_zero_g}.

\begin{figure*}[htbp]
    \centering
    \includegraphics[width=0.9\linewidth]{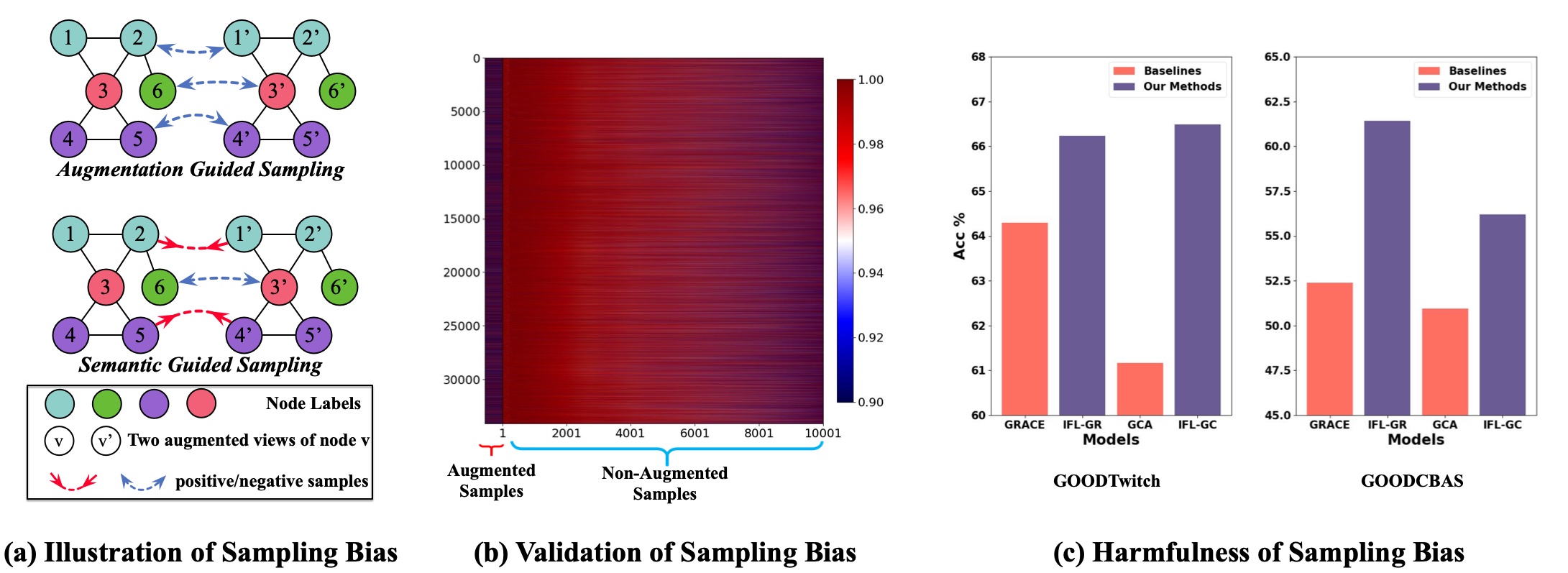}
    \caption{(a) illustrates the sampling bias via a case example; (b) validates sampling bias by comparing the nodes' representation similarity between augmented samples and top-20 non-augmented ones by a supervised graph encoder on GOODTwitch; (c) shows the harmfulness of sampling bias by comparing downstream task performances of traditional GCL baselines and our bias-corrected methods.}
    \label{fig:illustration_validation_harmfulness}
\end{figure*}

Generally, traditional GCL methods optimize InfoNCE loss \cite{InfoNCE} to define self-supervised tasks \cite{GRACE, GCA}. Specifically, they leverage graph data augmentation\cite{liu2022graph_aug, rong2019dropedge_aug} to generate different views of the graph and sample augmented pairs as positive samples and non-augmented pairs as negative samples. 
Returning to the core objective of GCL, it aims to bring samples with similar semantics closer together and push samples with dissimilar semantics farther apart\cite{wang2020understanding_alignment_uniformity}.
The aforementioned approach essentially uses augmentations as a proxy for semantics, assuming that augmented samples share similar semantic information and can be treated as positive samples, while non-augmented samples are considered negative.
However, detailed analysis and experimental validation reveal that traditional GCL methods suffer from sampling bias, limiting their performance \cite{liu2023b2_bias, zhao2021graph_bias}.
As shown in Fig.\ref{fig:illustration_validation_harmfulness} (a), in augmentation-guided sampling, non-augmented node pairs like \((node1, node2)\) and \((node4, node5)\) are treated as negative samples and pushed apart, despite their high semantic similarity (in terms of position, features, and structure). This prevents the model from correctly capturing semantic information in the original graph, leading to sampling bias and affecting downstream task performance.
To validate this, we compare the node representation similarity between augmented and non-augmented samples in the real dataset GOODTwitch, as shown in Fig.\ref{fig:illustration_validation_harmfulness} (b), where some non-augmented samples exhibit even higher similarity than augmented ones.
Furthermore, as shown in Fig.\ref{fig:illustration_validation_harmfulness} (c), this sampling bias harms GCL performance on downstream tasks.

To combat the above challenges, we argue that GCL is essentially a Positive-Unlabeled (PU) learning problem, where the training data consists of positive samples and unlabeled samples, with no explicit negative sample labels \cite{shift_PU, debiased_gcl_pu, PUCL, PUCL_1, pu_survey}.
It aligns with the actual situation in GCL i.e., augmented samples with similar semantics are considered as labeled positive samples, while others with unknown semantics are treated as unlabeled. 
From this perspective, the key of GCL lies in extracting semantic information and leveraging it to resample contrastive samples for the self-supervised task.

To achieve the above objective, we propose IFL-GCL where "IFL" means using \underline{\textbf{I}}nfoNCE as a "\underline{\textbf{F}}ree-\underline{\textbf{L}}unch"  to extract semantic information and resample contrastive samples. 
Specifically, we discover that the representation similarity trained with InfoNCE \cite{InfoNCE} shares the same order with the probability of the contrastive sample being positive through theoretical analysis. 
In other words, InfoNCE brings "free-lunch" enabling us to extract semantic information and resample contrastive samples via representation similarity. 
Based on these, we redefine the maximum likelihood objective of InfoNCE instead of heuristically modifying the samples in InfoNCE, and naturally derive a new InfoNCE loss which demonstrates theoretically superior bias correction capabilities.
Building on the above, IFL-GCL frames GCL within the PU learning framework, enabling semantically guided GCL, thereby mitigating sampling bias and enhancing GCL effectiveness.

To demonstrate the effectiveness of IFL-GCL, we validated its performance under two popular frameworks.
In the graph pre-training framework, IFL-GCL show significantly improvements in both IID and OOD scenarios with an accuracy improvement up to 9.05\% as shown in Fig.\ref{fig:illustration_validation_harmfulness} (c). 
When using LLM as an enhancer to porcess graph, IFL-GCL also brings consistent improvements showcasing its potential in graph foundation model research. 
In summary, our contributions are as follows:
\begin{itemize}
    \item \textbf{Promising Way:} We first introduce a GCL framework based on PU learning to achieve semantic guidance rather than augmentation-based guidance.
    \item \textbf{Innovative Method:} We propose IFL-GCL to use InfoNCE as a "free-lunch" to extract semantic information for resampling contrastive samples based on theoretical analysis, then redefine the maximum likelihood objective of InfoNCE and naturally derive a new InfoNCE loss function which exhibits a stronger bias correction capability.
    \item \textbf{Extensive Experiments:} Extensive experimental results on both graph pre-training and LLM as enhancers show significantly improvements of IFL-GCL under both IID and OOD scenarios, achieving up to a 9.05\% improvement, validating its effectiveness with semantical guidance.
\end{itemize}

\section{Preliminaries}\label{preliminary}
In this section, we present the formal process of graph contrastive learning (GCL) and validate the sampling bias in traditional GCL.

\subsection{Graph Contrastive Learning}
Due to the scarcity of labeled data, label-free self-supervised learning paradigms such as \cite{graph_ssl_survey_1, graph_ssl_survey_2, hu2020gpt_ssl, jin2020self_gssl, jing2021hdmi_gssl, rong2020self_gssl, peng2020self_gssl, wang2021self_gssl, you2021graph_gssl} have become mainstream in the field of graph machine learning, with graph contrastive learning being one of the most widely studied branches \cite{gcl_survey, BGRL,GRACE,GCA, GCC, MVGRL, DGI}. 

\textbf{Problem Formulation}. Let \(\mathbf{G}=(\mathbf{X, A})\) denote a graph, where \(\mathbf{X}\in \mathbb{R}^{N\times F}\) denotes the nodes' feature map, and \(\mathbf{x}_i\) is the feature of \(i\)-th node \(n_i\). \(\mathbf{A} \in \mathbb{R}^{N \times N}\) denotes the adjacency matrix, 
where \(\mathbf{A}_{ij} = 1\) if and only if there is an edge from \(n_i\) to \(n_j\). 
GCL aims at training a GNN encoder \(f_{\theta}(\mathbf{G})\) that maps graph \(\mathbf{G}\) into the node representations \(\mathbf{H}\in \mathbb{R}^{N \times d}\) in a low-dimensional space, which captures the essential intrinsic information from both features and structure.

\textbf{General Paradigm}. First, two augmented views, \({\mathbf{G}}^{aug_1} = aug_1(\mathbf{G})\) and \({\mathbf{G}}^{aug_2} = aug_2(\mathbf{G})\), are obtained from the original input graph through two sampled graph data augmentation techniques, such as feature masking, edge dropping, and others \cite{rong2019dropedge_aug, liu2022graph_aug}. 
Next, we sample the contrastive samples from these two augmented graphs. In this paper, we take the node-node level contrast \cite{DGI, GRACE, GCA} as an example: 
\begin{equation}\label{eq:aug_samples}
    \begin{aligned}
        D^{aug+}&=\{(u_i, v_i), (v_i, u_i)\}_{i=1}^N \\
        D^{aug-}&=\{(u_i, v_j), (v_i, u_j)\}_{i \neq j, i,j=1}^N
    \end{aligned}
\end{equation}
where \(D^{aug+}\) means the set of augmented pairs which are sampled as positive samples and \(D^{aug-}\) means the set of other non-augmented pairs which are sampled as negative ones;  \(u_i\), \(v_j\) means the \(i\)-th node in the \(\mathbf{G}^{aug_1}\) and \(j\)-th node in \(\mathbf{G}^{aug_2}\) correspondingly. It's worth noting that \((u_i, v_j)\) and \((v_i, u_j)\) are two different contrastive samples which have different augmented graph combination order for the \(i\)-th and \(j\)-th nodes in the original graph.
Finally, the contrastive loss function such as \cite{InfoNCE, dong2018triplet_contrastive_loss, sohn2016improved_contrastive_loss} is used to optimize the model, ensuring that the similarity between positive samples is high while the similarity between negative samples is low thereby learning feature representations with high discriminability that can quickly adapt to various downstream tasks. Take InfoNCE \cite{InfoNCE} loss for example:
    \begin{align}
        \label{eq:local_l} l_{u_i,v_i} &=-\log{\frac{s_{\theta}(u_i, v_i)}{\sum_{j\neq i,j=1}^{N} s_{\theta}(u_i, u_j) + \sum_{j=1}^{N}s_{\theta}(u_i,v_j)}} \\
        \label{eq:global_l} L &= \frac{1}{2N} \sum_{i=1}^N l_{u_i,v_i} + l_{v_i, u_i}
    \end{align}
where \(s_{\theta}(u_i, v_i)\) measures the representation similarity between node \(u_i\) and node \(v_j\) as follows:
\begin{equation}\label{eq:s_function_def}
    s_{\theta}(u_i, v_j) = \exp ( cos(\mathbf{U}_i, \mathbf{V}_j) / \tau )
\end{equation}
where \(\mathbf{U} = f_{\theta}(\mathbf{G}^{aug_1})\), \(\mathbf{V} = f_{\theta}(\mathbf{G}^{aug_2})\) and \(\mathbf{U}_i\) means the \(i\)-th node's representation in \(\mathbf{G}^{aug_1}\), \(\mathbf{V}_j\) means the \(j\)-th node's representation in \(\mathbf{G}^{aug_2}\);
$cos(\cdot, \cdot)$ is the cosine similarity function and $\tau$ is the temperature which controls the diameter of the representation space.
\(l_{u_i, v_i}\) is the local loss produced by contrastive sample \((u_i, v_i)\), \(L\) is the the final global loss which is the mean local loss of all samples of \(D^{aug+}\) shown in Eq.\ref{eq:aug_samples}.

\subsection{Validation of Sampling Bias in GCL}
\begin{figure}[htbp]
    \centering
    \includegraphics[width=0.8\linewidth]{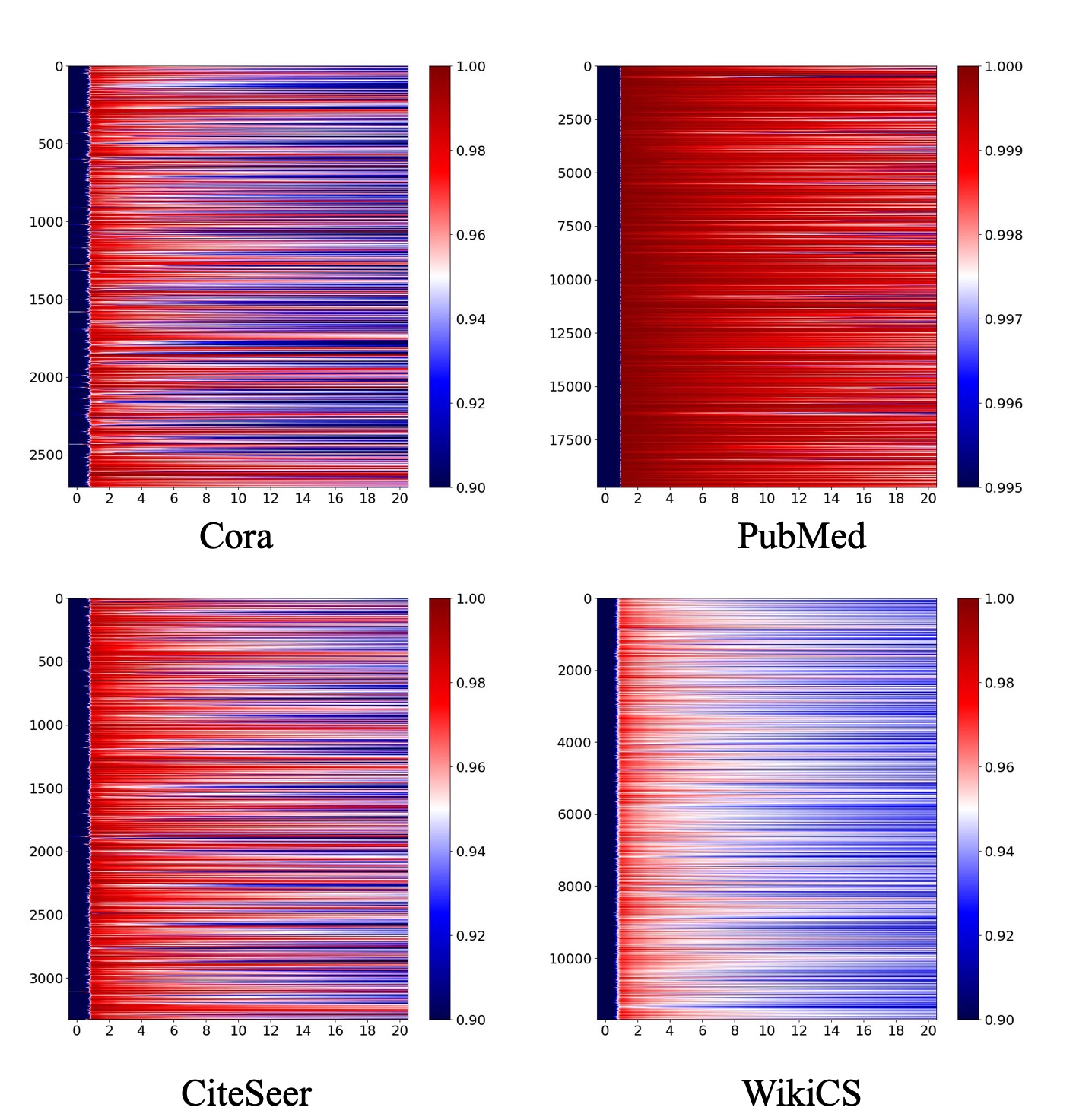}
    \caption{Semantics similarity matrix of $\mathbf{G}^{aug_1}$ and $\mathbf{G}^{aug_2}$ after supervised-learning which is rearranged with the $D^{aug+}$ at the first column and decently sorted $D^{aug-}$ as follows.}
    \label{fig:valid_fig}
\end{figure}
In this section, we consider the node representation similarity learned under a large amount of supervision as an approximation of semantic similarity, and use it to verify that there are semantic similar samples among non-augmented ones which are classified as negative and cause sampling bias. 

Specifically, we train a graph encoder $f_{\theta}(\cdot)$ using a supervised learning paradigm to capture the semantic information to the best extent:
\begin{equation}\label{eq:sup_train_GNN}
    \theta^*=\mathop{\arg\min}\limits_{\theta} L_{sup}(f_{\theta}(\mathbf{G}), Y)
\end{equation}
where \(f_{\theta}(\cdot)\) is the graph neural network encoder: GAT\cite{GAT}, \(L_{sup} \) is the supervised loss function for node classification, \(\mathbf{G}\) is the graph of real-word dataset: Cora, PubMed, CiteSeer or WikiCS \cite{cora, pubmed, citeseer, wikics} and \(Y\) is the nodes' label set. Due to the extensive use of supervisory signals, the trained encoder can model node semantic information effectively. It is worth noting that although the encoder used here is trained for a specific task, the method of capturing semantic information through supervisory signals is applicable to any task.

Then we use $f_{\theta^*}(\cdot)$ to capture the semantic information of the augmented graph in GCL:
\begin{equation*}
    \begin{aligned}
        \mathbf{U^*} = f_{\theta^*}(\mathbf{G}^{aug_1})\quad
        \mathbf{V^*} = f_{\theta^*}(\mathbf{G}^{aug_2})
    \end{aligned}
\end{equation*} 
We calculate the cosine similarity matrix \(\mathbf{S}\) of \(\mathbf{U^*}\) and \(\mathbf{V^*}\):
\begin{equation*}
    \begin{aligned}
        \mathbf{S} = \frac{\mathbf{U^*} \cdot \mathbf{V^*}^T }{\|\mathbf{U^*}\| \cdot \|\mathbf{V^*}\|^T}
    \end{aligned}
\end{equation*}
where \(\|\mathbf{U^*}\|\) is an \(N \times 1\) vector containing the \(L2\) norm of each row of matrix \(\mathbf{U^*}\);
$\mathbf{S}_{ij}$ means the cosine similarity between $\mathbf{U^*}_i$ and $\mathbf{V^*}_j$. 
For the convenience of observation, we rearranges $\mathbf{S}$ to get $\mathbf{S'}$ by extracting its main diagonal as the first column and sorting the remaining columns in descending order row by row:
\begin{equation}
    \mathbf{S'} = \begin{bmatrix}
        \mathbf{S}_{11} & sorted(\mathbf{S}_{12}, \mathbf{S}_{13}, \cdots, \mathbf{S}_{1N}) \\
        \mathbf{S}_{22} & sorted(\mathbf{S}_{21}, \mathbf{S}_{23}, \cdots, \mathbf{S}_{2N} ) \\
        \vdots & \vdots \\
        \mathbf{S}_{NN} & sorted(\mathbf{S}_{N1}, \mathbf{S}_{N2}, \cdots, \mathbf{S}_{NN-1}) 
    \end{bmatrix}
\end{equation}

The rearranged similarity matrix with first column and top-20 remaining columns \(\mathbf{S'}_{:,0:20} \) is extracted and shown in Fig.\ref{fig:valid_fig}. 
The first column in \(\mathbf{S'}\)  shows the cosine similarity between nodes in augmented samples \(D^{aug+}\) and others corresponds to the top-20 most semantically similar ones in non-augmented samples \(D^{aug-}\). 
From Fig.\ref{fig:valid_fig}, we observe that the first columns of the matrix exhibit a dark blue color, while the subsequent columns transition from red to blue. From such a distribution, it can be inferred that some node pairs in \(D^{aug-} \)exhibit semantic similarity either higher than (the red ones) or close to ( the blue ones) that of the samples in \(D^{aug+} \), proving our perspective that GCL with augmentation-guided sampling approach suffers from sampling bias because of the misclassified semantic similar non-augmented samples.

\section{Methodology} 
In this section we propose IFL-GCL which uses \underline{\textbf{I}}nfoNCE as a "\underline{\textbf{F}}ree-\underline{\textbf{L}}unch" to extract semantic information for resampling the contrastive samples, in order to correct the sampling bias in traditional GCL.
We elaborate on the following topics: the first topic explains our motivation that GCL is a Positive-Unlabeled learning problem and clarify that the root cause of the sampling bias is semantically similar non-augmented samples. 
Then we focus on extracting semantic information and updating the training objective to correct the sampling bias. 
The third part discusses the theoretical superiority of our proposed method IFL-GCL.

\subsection{Motivation}
We point out that GCL is essentially a Positive-Unlabeled (PU) learning problem.
We define two flag signs: semantic label \(y=\{-1, +1\}\) and labeling status \(o=\{-1,+1\}\), where \(y(\mathbf{x}) = +1\) means \(\mathbf{x}\) belongs to positive class and \(y(\mathbf{x})=-1\) means it's negative; \(o(\mathbf{x})=+1\) means \(\mathbf{x}\) is labeled and vice versa. 
It is worth noting that \(y\) reflects the confirmed semantic properties of the data, while \(o\) just reflects samples' labeling status with unknown semantic information.

Traditional GCL methods substitute positive/negative samples with augmented/non-augmented ones:
\begin{equation}
    \label{eq:pos/neg=aug+/-}
    \begin{aligned}
        D^+ &= D^{aug+} \\
        D^- &= D^{aug-}
    \end{aligned}
\end{equation}
where \(D^+ = \{\mathbf{x}|y=+1\}\), \(D^- = \{\mathbf{x}|y=-1\}\).
In fact, augmented samples have confirmed semantically similarity and the semantic information of non-augmented samples remains unknown, therefore they should be viewed as follows:
\begin{equation}\label{eq:aug+_L+,aug-_U}
    \begin{aligned}
        D^{aug+} &= D_L^+  \\
        D^{aug-} &= D_U = D_U^+ \cup D_U^-
    \end{aligned}
\end{equation}
where \(D_L^+ = \{\mathbf{x} | y= +1, o = +1  \} \) ,\(D_U = \{\mathbf{x} | o = -1\}\) and the unlabeled samples can further be divided into unlabeled positive samples \(D_U^+ = \{\mathbf{x} | y=+1, o = -1\}\) and unlabeled negative samples \(D_U^-= \{\mathbf{x} | y=-1, o = -1\}\) according to their semantics.
PU learning focus on the tasks where data are split into positive labeled samples and unlabeled ones which is more aligned with the contrastive samples in traditional GCL.
Therefore we treat GCL as a PU learning problem and reconsider the data composition as the following equations:
\begin{equation}\label{eq:correct_by_U+}
    \begin{aligned}
        D^+ &= D_L^+ \cup D_U^+ \\
        D^- &= D_U - D_U^-
    \end{aligned}
\end{equation}
Viewing GCL as a PU learning problem, it's evident that the semantic similar non-augmented samples \(D_U^+\) are misclassified as negative samples in traditional GCL which leads to the sampling bias.

\subsection{Method}
In this section, we propose IFL-GCL which uses infoNCE as a free-lunch to extract semantic information for resampling and update the training objective based on the resampling results. 

\subsubsection{\textbf{Extracting Semantic Information to Resample}}
Firstly, we demonstrate the Invariance of Order (IOD) assumption and its corollary, introducing the key probability density ratio function \(r(x)\). Next, we proof that InfoNCE in GCL exactly models \(r(x)\). Finally, through \(r(x)\) and the corollary of the Invariance of Order assumption, we obtain a classifier that allows us to extract \(D_U^+\) for correcting the sampling bias.

\textbf{Invariance of Order Assumption and Corollary.}
Kato et al. \cite{IOD_assuption}
proposed a relaxed and universal assumption known as the \textbf{I}nvariance of \textbf{O}r\textbf{D}er (\textbf{IOD}) as follows: 
\begin{equation}\label{eq:IOD_assumption}
    \begin{aligned}
        &\forall \mathbf{x},\mathbf{\hat{x}}  \in D:  \\
        &p(y=+1|\mathbf{x}) \leq p(y=+1|\mathbf{\hat{x}}) \Leftrightarrow  p(o=+1|\mathbf{x}) \leq p(o=+1|\mathbf{\hat{x}})
    \end{aligned}
\end{equation}
which means that the higher probability of a contrastive sample belongs to the positive class, the greater the likelihood that it will be labeled as positive, and vice versa. 
Kato et al. also derived the following significant corollary under the IOD assumption:
\begin{equation}
    \begin{aligned}\label{eq:IOD_corollary}
        &\forall \mathbf{x},\mathbf{\hat{x}}  \in D:  \\
        &p(y=+1|\mathbf{x}) \leq p(y=+1|\mathbf{\hat{x}}) \Leftrightarrow  
        r(\mathbf{x}) \leq r(\mathbf{\hat{x}})
    \end{aligned}
\end{equation}
where $r(\mathbf{x})$ is the distribution density ratio as following:
\begin{equation}\label{eq:r(x)}
    r(\mathbf{x}) := \frac{p(\mathbf{x} | y=+1,o=+1)}{p(\mathbf{x})}
\end{equation}
According to Eq.\ref{eq:IOD_corollary} and Eq.\ref{eq:r(x)}, the of density ratio between data $D_L^+=\{\mathbf{x}|y=+1, o=+1\}$ and $D=\{\mathbf{x}\}$ reflects the order of the positive class posterior probability \(p(y=+1|\mathbf{x})\). 
Therefore, we can derive a classifier $h(\cdot)$ which can be used to find semantically similar samples after modeling the $r(\cdot)$ and given a threshold $t_r$ as follows:
\begin{equation}\label{eq:classifier_based_on_r(x)}
    y(\mathbf{x}) = h(\mathbf{x}):= sign(r(\mathbf{x}) - t_r)
\end{equation}
where \(y(\mathbf{x})\) is the class label \(y\) of contrastive sample \(\mathbf{x}\).

\textbf{Free-Lunch Provided by InfoNCE.}
Instead of employing traditional density ratio estimation methods such as LSIFs\cite{LSIF, uLSIF, RuLSIF} to model \(r(\mathbf{x})\), we astutely note that the InfoNCE loss function used in Eq.\ref{eq:local_l} inherently provides a "free-lunch": the representations similarity trained under InfoNCE is proportional  to the density ratio \(r(\mathbf{x})\). We will now proof this "free-lunch".

Firstly, we notice that the initial design of InfoNCE in \cite{InfoNCE} was motivated by the aim to model a density ratio as follows: 
\begin{equation}\label{eq:InfoNCE_motivation_model}
    f_k(x_{t+k} | c_t) \propto \frac{p(x_{t+k}|c_t)}{p(x_{t+k})}
\end{equation}
where \(\propto\) denotes "proportional to", \(x_{t+k}\) is the future observations, \(c_t\) is the anchored sample and \(f_k(\cdot, \cdot)\) is the model which targets at preserves the mutual information between \(x_{t+k}\) and \(c_t\). 
Under the context of GCL, the \(f_k(\cdot, \cdot)\) corresponds to \( s_{\theta}(\cdot, \cdot) \) in Eq.\ref{eq:s_function_def}.
Generally speaking, given a node pair \((n,n')\) where $n$ serve as the anchor, the \(s_{\theta}(\cdot, \cdot)\) learned by GCL under InfoNCE essentially models a nodes' probability density ratio as follows:
\begin{equation} \label{eq:infoNCE_propto_ratio}
    s_{\theta}(n, n') \propto \frac{q(n | n')}{q(n)} = \frac{q(n, n')/q(n')}{q(n)} = \frac{q(n,n')}{q(n)q(n')}
\end{equation}
where \(q(\cdot)\) is the probability density of node. 

Next, we connect the contrastive samples' distributions and the nodes' distributions. 
Considering the labeled positive contrastive samples \(\mathbf{x} = (n, n')\) with \(y=+1, o=+1\) which is used to train \(s_{\theta}(\cdot, \cdot)\), they are \textit{\textbf{non-independent}} pairs of nodes connected together by data augmentation.  
Thus the distribution of labeled positive contrastive samples can be represented as the joint distribution of the node pair:
\begin{equation}\label{eq:pL+(x)}
    p(\mathbf{x}=(n,n') | y=+1, o=+1) = q(n, n')
\end{equation}
Conversely, if a contrastive sample \(\mathbf{x} = (n, n')\) is constructed without any information regarding \(y\) and \(o\), it is obtained by \textit{\textbf{independently}} sampling two nodes \(n\) and \(n'\), which can be expressed as:
\begin{equation}\label{eq:p(x)}
    p(\mathbf{x}=(n, n'))=q(n)q(n')
\end{equation}
By combining Eq.\ref{eq:p(x)}, Eq.\ref{eq:pL+(x)} and Eq.\ref{eq:r(x)}, we obtain:
\begin{equation}\label{eq:s_eq_r}
    s_{\theta}(n,n') \propto \frac{q(n,n')}{q(n)q(n')}=\frac{p(\mathbf{x}|y=+1, o=+1)}{p(\mathbf{x})} = r(\mathbf{x})
\end{equation}
which means that \(s_{\theta}(n,n')\) is proportional to \(r(\mathbf{x})\).
Finally, combining the conclusion derived from Eq.\ref{eq:IOD_corollary}, we complete the proof of infoNCE's free-lunch as shown below: 
\begin{equation}\label{eq:Free-Lunch}
    \begin{aligned}
        &\forall \mathbf{x}=(n,n'), \mathbf{\hat{x}}=(\hat{n},\hat{n}') \in D: \\
        &p(y=+1|\mathbf{x}) \leq p(y=+1|\mathbf{\hat{x}}) \Leftrightarrow 
        s_{\theta}(n,n') \leq s_{\theta}(\hat{n}, \hat{n}')
    \end{aligned}
\end{equation}
which means that training with infoNCE not only optimizes the contrastive loss but also provides a "free-lunch" to model the positive class posterior probability \(p(y=+1|\mathbf{x})\). 

This free-lunch allows us to identify the $D_U^+$ samples like Eq.\ref{eq:classifier_based_on_r(x)} with a new threshold $t_s$ corresponds to \(s_{\theta}(\cdot, \cdot)\) as hyper-parameter:
\begin{equation}\label{eq:classifier_based_on_s}
    y(\mathbf{x}) = h_s(\mathbf{x}=(n,n');\theta):=sign(s_{\theta}(n,n')-t_s)
\end{equation}
By this point, we can extract semantically similar non-augmented samples \(D_U+\) from \(D_U\) for resampling based on Eq.\ref{eq:correct_by_U+}:
\begin{equation}\label{eq:resample}
    \begin{aligned}
    &D_U^+ = \{(n,n')\}_{sign(s_{\theta}(n,n')-t_s)=+1}^{(n,n') \in D_U} \\
        &D^+ = D_L^+ \cup D_U^+ \\
        &D^- = D_U - D_U^+\\
    \end{aligned}
\end{equation}
which means that the semantically similar non-augmented samples \(D_U+\) is treated as positive samples instead of negative ones.

\subsubsection{\textbf{Updating Training Objective}}
After resampling, we introduce \(D_U^+\) into the GCL training process, ensuring the correction of sampling bias. 
We point out that the Eq.\ref{eq:local_l} can be understood as the form of negative-log-likelihood probability as shown below: 
\begin{equation}\label{local_J_as_NegLogP}
    J_{u_i,v_i} :=-\log{\frac{s_{\theta}(u_i, v_i)}{\sum_{j\neq i,j=1}^{N} s_{\theta}(u_i, u_j) + \sum_{j=1}^{N}s_{\theta}(u_i,v_j)}} = -\log \mathbf{P}_{u_i,v_i}  
\end{equation}
where $\mathbf{P}_{u_i,v_i}$ is the normalized probability of numerator $s_{\theta}(u_i, v_i)$ over all possible $v_j$ given $u_i$ as an anchor node. 
Furthermore, we point out that \(\mathbf{P}_{n,n'}\) can be regarded as  the approximation of likelihood probability of a general node pair $(n, n')$ being positive contrastive sample when $n$ is the anchor sample. According to Eq.\ref{eq:Free-Lunch},  \(s_{\theta}(n,n')\) satisfies the order-invariant relationship with the probability of contrastive sample \((n,n')\) being positive. Note that when \( n \) is given as the anchor node, the denominator of the normalization for \( s_{\theta}(n,n') \) to obtain \( \mathbf{P}(n,n') \) is consistent. 
For example, given \(u_i\) as anchor node, for any \(v_j\) the denominator of \(\mathbf{P}_{u_i,v_j}\) is always \(\sum_{j\neq i,j=1}^{N} s_{\theta}(u_i, u_j) + \sum_{j=1}^{N}s_{\theta}(u_i,v_j) \). Therefore, based on Eq.\ref{eq:Free-Lunch}, the order of \( \mathbf{P}_{n,n'} \) and \( p(y = +1| (n, n')) \) is consistent when node \( n \) serves as the anchor:
\begin{equation}\label{P_is_likelyhood}
    \begin{aligned}
        & \forall n', n'' \in \mathcal{N}: \\
        & \mathbf{P}_{n,n'} \leq \mathbf{P}_{n,n''} \Leftrightarrow 
        p(y=+1|(n,n')) \leq p(y=+1|(n,n''))
    \end{aligned}
\end{equation}
where \(\mathcal{N}\) is the set of nodes. 
Thus, the normalized \( \mathbf{P}_{n,n'} \) can be seen as an approximation of \( p(y=+1|(n,n')) \).

Under this understanding, \(L\) in Eq.\ref{eq:global_l} can be naturally interpreted as the expectation of negative-log-likelihood probabilities of all labeled-positive samples:
\begin{equation}
    \begin{aligned}
        L:=& \frac{1}{N}\sum_{i=1}^N\frac{J_{u_i,v_i}+J_{v_i,u_i}}{2} \\
        =& \mathbb{E}J_{n,n'} \\
        =& \mathbb{E} -\log{\mathbf{P}_{n,n'}} 
    \end{aligned}
\end{equation}
where \((n,n') \in D_L^+ =  D^{aug+} = \{(u_i,v_i),(v_i,u_i)\}_{i=1}^N\).
Therefore, minimizing the InfoNCE loss function is equivalent to maximizing the likelihood probability of all positive samples, i.e., \(D_L^+\) in traditional GCL from the perspective of PU learning. 

After resampling in Eq.\ref{eq:resample}, the scope of positive samples is expanded to \(D_L^+ \cup D_U^+\), therefore, the likelihood probability of \(D_U^+\) must also be incorporated into the optimization objective: 
\begin{align}
    \label{eq:corrected_local_L} 
    L_{n, n'}^{corrected}&:= 
    -\log (\mathbf{P}_{n, n'} \prod_{(n, n'') \in D_U+}(\mathbf{P}_{n, n''})^{\beta \hat{s}_{\theta}(n, n'')})  
\end{align}
As shown in the following Eq.\ref{eq:corrected_local_L}, considering that our method has different confidence for different \(D_U^+\) samples, we assign exponential weights \(\hat{s}_{\theta}(n,n'')\) to the negative-log-likelihood probabilities of each \(D_U^+\) sample \((n,n'')\). The weight is the globally normalized similarity score as follows: 
\begin{equation}
    \hat{s}_{\theta}(n, n') = \frac{
    s_{\theta}(n, n') - \min \{s_{\theta}(n_i, n'_j)\}_{i,j =1}^{N}
    }
    {\max \{s_{\theta}(n_i, n'_j)\}_{i,j =1}^{N}}
\end{equation}
The more similar of \((n, n')\), the higher weight of \(\mathbf{P}_{n, n'}\). 
Moreover, considering the differences between \(D_L^+\) and \(D_U^+\), we further assign an exponential weight \(\beta\) to the negative-log-likelihood probabilities of all \(D_U^+\) samples comparing to the weight 1 for \(D_L^+\) samples.  After expanding the loss based on a certain \(D_L^+\) sample, taking the expectation will yield the final loss:
\begin{equation}
    \begin{aligned}
        \label{eq:corrected_local_L} L^{corrected} :=& \mathop{\mathbb{E}}\limits_{(n,n')\in D_L^+} L_{n,n'}^{corrected} \\
        =& \mathop{\mathbb{E}}\limits_{
            \substack{
            (n,n')\in D_L^+ \\ 
            (n,n'')\in D_U^+
            }
        } -log (\mathbf{P}_{n, n'}\prod(\mathbf{P}_{n, n''})^{\beta \hat{s}_{\theta}(n, n'')})
    \end{aligned}
\end{equation}

To mitigate the cumulative negative impact of individual erroneous \(D_U^+\) instances, we employ a dynamic updating method for \(D_U^+\). Specifically, we first train the model for \(M\) epochs using Eq.\ref{eq:global_l}. Then, every \(K\) training epochs, we use the representation similarity from the model obtained in the previous stage to resample the \( D_U^+ \), and update the loss function for training the graph encoder. This process repeats until the best checkpoint of our graph encoder \(\theta_{corrected}^*\). The pseudo-code of the whole pre-training process  is shown in Algorithm 1. 
\begin{algorithm}
  \caption{The whole pre-training process of our semantic-guided sampling approach for GCL}
  \label{algorithm:1}
  \begin{algorithmic}
      \Require origin Graph \(\mathbf{G}\), augmentation functions \(aug_1(\cdot)\) and \(aug_2(\cdot)\), randomly initialized graph encoder \(theta\), warm-up epochs \(M\), update interval epochs \(K\), threshold \(t_s\), max updating objective times \(T\), learning rate \(\alpha\)
      \Ensure optimized graph encoder \(\theta^*_{corrected}\)
      
      \State \(\mathbf{G}^{aug_1} \gets aug_1(\mathbf{G})\), \(\mathbf{G}^{aug_2} \gets aug_2(\mathbf{G})\)     
      \State \(D^+ \gets D^{aug+}\), \(D^- \gets D^{aug-}\)\Comment{Eq.\eqref{eq:aug_samples}}

      \State \(\theta^{(0)} \gets \theta\)
      \State \(L^{(0)} \gets L(\theta^{(0)}) \)  \Comment{Eq.\eqref{eq:global_l} }
      
    \For{\(i = 1\) \textbf{to} \(M\)}
        \State $\theta^{(0)} \gets \theta^{(0)} - \alpha \cdot \nabla_{\theta} L^{(0)}$ \Comment{Warming Up \(\theta^{(0)}\)}
    \EndFor

    \State \(\theta^{(1)} \gets \theta^{(0)} \)
    \For{\(t = 1, \ldots, T\)}
        \State \(h^{(t)}(n,n';\theta^{(t)}) \gets sign(s_{\theta^{(t)}}(n,n')-t_s)\) \Comment{Free-Lunch}
        \State \(D_U^{+, (t)} \gets \{(n, n')\}_{h^{(t)}(n, n'; \theta^{(t)}) = 1}\)
        \State \(D^+ \gets D^{aug+} \cup D_U^{+, (t)}\)
        \State \(D^- \gets D^{aug-} - D_U^{+, (t)}\) \Comment{Resample}
        \State \(L^{(t)} \gets L^{corrected}(\theta^{(t)})\)\Comment{Eq.\eqref{eq:corrected_local_L} }
        \For{\(i = 1\) \textbf{to} \(K\)}
            \State \(\theta^{(t)} \gets \theta^{(t)} - \alpha \cdot \nabla_{\theta} L^{(t)}\) \Comment{Update Training Objective}
        \EndFor
    \EndFor

    \State \textbf{Return} \(\theta^*_{corrected} \gets \theta^{(T)}\)
  \end{algorithmic}
  
\end{algorithm}

\subsection{Discussion}
This section discusses the theoretical advantages of our method compared to other approaches such as \cite{PUCL, PUCL_1, shift_PU, 
 debiased_gcl_pu}. In summary, our advantages are twofold: First, the assumptions we use to identify \( D_U^+ \) through the free lunch of InfoNCE are more aligned with the actual conditions in GCL. Second, by updating the training loss based on the maximum likelihood objective of InfoNCE, we impose a stronger constraint on \( D_U^+ \), resulting in more thorough bias correction.

\subsubsection{\textbf{General Assumption for GCL}}
Generally, PU learning methods adopt the \textbf{S}elect \textbf{C}ompletely \textbf{A}t \textbf{R}andom (\textbf{SCAR}) assumption \cite{pu_survey} which means the labeled positive samples are selected completely at random from the positive distribution independent from the sample's attributes. 
Under the SCAR assumption, both \(D_L^+\) and \(D_U^+\) are considered to be independently and identically distributed as the overall positive data \(D^+ = D_L^+ \cup D_U^+\). Consequently, it is tempting to conclude that \(D_L^+\) and \(D_U^+\) share the same distribution.
However, the SCAR assumption is too strong to hold because \(D_L^+\) in GCL is constructed artificially via data augmentation while \(D_U^+\) are the ones sharing similar position, features and structure information that exits inherently in the original graph. 
Unlike the SCAR assumption, the IOD assumption used in IFL-GCL is more relaxed and aligns more closely with the practical scenarios of GCL. 
For any contrastive sample \(\mathbf{x}=(n,n')\), if node \(n\) and \(n'\) is semantically similar, then the probability of \(\mathbf{x}\) being observed as a positive one is higher and vice versa.

\subsubsection{\textbf{Theoretically Solid Corrected Objective}}
Focus on the incorporating of newly discovered samples \(D_U^+\), we find that some other works  such as \cite{NaP, PUCL} .etc just heuristically modify the InfoNCE loss function without understanding the original objective from the perspective of maximum likelihood.
Specifically, to put the newly found \(D_U^+\) in model's training, these works can be approximated as employing a linear combination of the likelihoods of \(D_L^+\) and \(D_U^+\)  as shown below:
\begin{equation}\label{eq:other_corrected_J}
    \begin{aligned}
        l_{u_i,v_i}^{corrected} &= -\log{
        \frac{
        s_{\theta}(u_i, v_i) + \sum_{i,j=1}^N \alpha_{ij} s_{\theta}(u_i, v_j)
        }{
        \sum_{j\neq i,j=1}^{N} s_{\theta}(u_i, u_j) + \sum_{j=1}^{N}s_{\theta}(u_i,v_j)
        }
    } \\
    &= -log (\mathbf{P}_{u_i,v_i} + \sum_{i,j=1}^N \alpha_{ij} \mathbf{P}_{u_i, v_j} )
    \end{aligned}
\end{equation}
where \(\mathbf{P}_{u_i,v_i}\) and \(\mathbf{P}_{u_i, v_j}\) corresponds to the likelihood of \(D_L^+\) and \(D_U^+\).
Optimizing this loss is equivalent to maximizing a linearly combined likelihood of \(D_L^+\) and \(D_U^+\). 
It only requires a sufficiently large \(\mathbf{P}{u_i,v_i}\) for \(D_L^+\), imposing no strong constraints on the likelihood of \(D_U^+\). 
In contrast, our proposed loss in Eq.\ref{eq:corrected_local_L} adopts an exponential weighted product of the likelihoods of both \(D_U^+\) and \(D_L^+\). Our loss is more affected and sensitive to the likelihood of \(D_U^+\), which results in a more rigorous constraint on \(D_U^+\), leading to stronger bias correction.

\section{Experiments}
\begin{table*}[htbp]
    \centering
    \caption{Comparing our methods with 7 baselines across 9 datasets. We show the node classification accuracy, along with the corresponding standard deviation in parentheses. The best performance achieved on each dataset is highlighted in \underline{\textbf{bold underlined}}, and the second-best performance is indicated as \textit{\textbf{bold italic}}. We also show the improvement of our method compared to the directly related methods in the last two rows where \(\mathbf{\Delta_{GR}}\) represents the increase of IFL-GR compared to GRACE, and \(\mathbf{\Delta_{GC}}\) denotes the increase of IFL-GC compared to GCA. }
    \renewcommand{\arraystretch}{1.2} 
    \setlength{\tabcolsep}{3.5pt} 
    \begin{tabular}{@{}ccccccc||ccc@{}} 

    \hline
    & \large{\textbf{Cora}} & \large{\textbf{PubMed}} & \large{\textbf{CiteSeer}} & \large{\textbf{WikiCS}}  & \large{\textbf{Computers}} & \multicolumn{1}{c||}{\large{\textbf{Photo}}} & \large{\textbf{GOODTwitch}} & \large{\textbf{GOODCora}} & \large{\textbf{GOODCBAS}} \\ 
    \midrule

    \makecell[c]{\large{\textbf{DGI}}\\}
    &\makecell[c]{\(82.99_{(1.38)}\)}
    &\makecell[c]{\(84.89_{(0.05)}\)}
    &\makecell[c]{\(69.79_{(0.40)}\)}
    &\makecell[c]{\(78.54_{(0.85)}\)}
    &\makecell[c]{\(86.97_{(0.07)}\)}
    &\makecell[c]{\(91.73_{(0.11)}\)}
    &\makecell[c]{\(59.67_{(2.26)}\)}
    &\makecell[c]{\(44.73_{(0.79)}\)}
    &\makecell[c]{\(57.62_{(1.78)}\)}\\

    \makecell[c]{\large{\textbf{COSTA}}\\}
    &\makecell[c]{\(84.61_{(0.28)}\)}
    &\makecell[c]{\(86.01_{(0.28)}\)}
    &\makecell[c]{\(71.31_{(0.62)}\)}
    &\makecell[c]{\(78.93_{(0.92)}\)}
    &\makecell[c]{\(88.56_{(0.40)}\)}
    &\makecell[c]{\(92.30_{(0.20)}\)}
    &\makecell[c]{\(62.12_{(0.99)}\)}
    &\makecell[c]{\(49.31_{(1.08)}\)}
    &\makecell[c]{\(41.90_{(3.75)}\)}\\

    \makecell[c]{\large{\textbf{BGRL}}\\}
    &\makecell[c]{ \(78.48_{(0.54)}\)}
    &\makecell[c]{ \(84.41_{(0.08)}\)}
    &\makecell[c]{ \(63.62_{(1.28)}\)}
    &\makecell[c]{ \(77.65_{(0.67)}\)}
    &\makecell[c]{ \(85.94_{(0.51)}\)}
    &\makecell[c]{ \(91.70_{(0.55)}\)}
    &\makecell[c]{ \(56.85_{(4.31)}\)}
    &\makecell[c]{ \(43.17_{(0.93)}\)}
    &\makecell[c]{ \(52.38_{(4.42)}\)}\\

    \makecell[c]{\large{\textbf{MVGRL}}\\}
    &\makecell[c]{ \(83.60_{(0.65)}\)}
    &\makecell[c]{ \(84.29_{(0.16)}\)}
    &\makecell[c]{ \(69.95_{(0.34)}\)}
    &\makecell[c]{ \(76.03_{(0.52)}\)}
    &\makecell[c]{ \(84.64_{(0.38)}\)}
    &\makecell[c]{ \(89.67_{(0.14)}\)}
    &\makecell[c]{ \(57.75_{(2.21)}\)}
    &\makecell[c]{ \(28.01_{(0.60)}\)}
    &\makecell[c]{ \(47.62_{(1.78)}\)}\\ 
    
    \makecell[c]{\large{\textbf{GBT}}\\}
    &\makecell[c]{ \(83.51_{(0.62)}\)}
    &\makecell[c]{ \(85.90_{(0.03)}\)}
    &\makecell[c]{ \(70.07_{(1.15)}\)}
    &\makecell[c]{ \(78.91_{(0.52)}\)}
    &\makecell[c]{ \(\underline{\textbf{89.34}}_{(0.09)}\)}
    &\makecell[c]{ \(92.93_{(0.10)}\)}
    &\makecell[c]{ \(59.37_{(2.31)}\)}
    &\makecell[c]{ \(47.22_{(0.40)}\)}
    &\makecell[c]{ \(52.38_{(4.42)}\)}\\

    \midrule
    
    \makecell[c]{\large{\textbf{GRACE}}\\}
    &\makecell[c]{\(83.97_{(0.38)}\)}
    &\makecell[c]{\(86.34_{(0.06)}\)}
    &\makecell[c]{\(70.75_{(0.96)}\)}
    &\makecell[c]{\(78.90_{(0.56)}\)}
    &\makecell[c]{\(88.51_{(0.36)}\)}
    &\makecell[c]{\(92.33_{(0.53)}\)}
    &\makecell[c]{\(64.29_{(1.25)}\)}
    &\makecell[c]{\(50.30_{(1.19)}\)}
    &\makecell[c]{\(52.38_{(5.99)}\)}\\

   \makecell[t]{\large{\textbf{GCA}}\\} 
    &\makecell[c]{\(84.79_{(0.37)}\)}
    &\makecell[c]{\(\textit{\textbf{87.13}}_{(0.16)}\)}
    &\makecell[c]{ \(70.57_{(1.31)}\)}
    &\makecell[c]{ \(77.99_{(0.60)}\)}
    &\makecell[c]{ \(88.63_{(0.42)}\)}
    &\makecell[c]{ \(\textit{\textbf{92.81}}_{(0.12)}\)}
    &\makecell[c]{ \(61.17_{(0.87)}\)}
    &\makecell[c]{ \(51.66_{(1.44)}\)}
    &\makecell[c]{ \(50.95_{(1.78)}\)}\\

    \midrule

    \makecell[c]{\large{\textbf{IFL-GR}}\\}
    &\makecell[c]{\(\underline{\textbf{85.40}}_{(0.42)}\)}
    &\makecell[c]{\(86.54_{(0.09)}\)}
    &\makecell[c]{\(\underline{\textbf{71.83}}_{(0.52)}\)}
    &\makecell[c]{\(\underline{\textbf{79.36}}_{(0.51)}\)}
    &\makecell[c]{\(89.00_{(0.08)}\)}
    &\makecell[c]{\(92.63_{(0.21)}\)}
    &\makecell[c]{\(\textit{\textbf{66.23}}_{(0.71)}\)}
    &\makecell[c]{\(\textit{\textbf{51.84}}_{(0.87)}\)}
    &\makecell[c]{\(\underline{\textbf{61.43}}_{(4.04)}\)}\\

\makecell[c]{\large{\textbf{IFL-GC}}\\}
    &\makecell[c]{\(\textit{\textbf{85.35}}_{(1.42)}\)} 
    &\makecell[c]{\(\underline{\textbf{87.37}}_{(0.08)}\)}
    &\makecell[c]{\(\textit{\textbf{71.35}}_{(0.81)}\)}
    &\makecell[c]{\(\textit{\textbf{79.22}}_{(0.56)}\)}
    &\makecell[c]{\(\textit{\textbf{89.33}}_{(0.21)}\)}
    &\makecell[c]{\(\underline{\textbf{93.13}}_{(0.25)}\)}
    &\makecell[c]{\(\underline{\textbf{66.49}}_{(1.09)}\)}
    &\makecell[c]{\(\underline{\textbf{52.55}}_{(0.11)}\)}
    &\makecell[c]{\(\textit{\textbf{56.19}}_{(1.78)}\)}\\
    
    \midrule

    \makecell[c]{\large{
    \(\mathbf{\Delta_{GR}}\)
    }\\}
    &\makecell[c]{+1.43\%}
    &\makecell[c]{+0.20\%}
    &\makecell[c]{+1.08\%}
    &\makecell[c]{+0.46\%}
    &\makecell[c]{+0.49\%}
    &\makecell[c]{+0.30\%}
    &\makecell[c]{+1.94\%}
    &\makecell[c]{+1.54\%}
    &\makecell[c]{+9.05\%}\\
    
   \makecell[c]{\large{
    \(\mathbf{\Delta_{GC}}\)
    }\\}
    &\makecell[c]{+0.57\%}
    &\makecell[c]{+0.24\%}
    &\makecell[c]{+0.78\%}
    &\makecell[c]{+1.23\%}
    &\makecell[c]{+0.71\%}
    &\makecell[c]{+0.32\%}
    &\makecell[c]{+5.32\%}
    &\makecell[c]{+0.89\%}
    &\makecell[c]{+5.24\%}\\
    \bottomrule
\end{tabular}
    \label{tab:Main Experiments}
\end{table*}
In this section, we perform thorough experiments to evaluate the effectiveness of our proposed method IFL-GCL. 
Under both paradigms of graph pre-training and LLMs as feature enhancement, we demonstrate the performance improvement of IFL-GCL over 7 traditional GCL baselines on 9 datasets for downstream node classification task. 
Additionally, we have conducted extensive empirical experiments, including: analyzing the semantics of \(D_U^+\) found by IFL-GCL, and examining the impact of three important hyper-parameters.

\subsection{\textbf{Experimental Settings }}

\textbf{Dataset Description.}
We evaluate our IFL-GCL on node classification task across 9 datasets under independent and identically distributed (IID) and out-of-distribution (OOD) scenarios.
We employ five commonly used graph datasets Cora, PubMed, CiteSeer, WikiCS, Computers, and Photo \cite{cora, pubmed, citeseer, wikics} for IID scenario where the train set and the test set exhibit similar distributions. 
We conduct experiments on three datasets from the GOOD benchmark \cite{GOOD}: GOODTwitch, GOODCora, and GOODCBAS, which are designed for OOD scenarios. 

\textbf{Baselines.} We use 7 generally used GCL methods as baselines, among which  MVGRL, GBT, BGRL, COSTA, DGI\cite{MVGRL, GBT, BGRL, costa, DGI} are not directly related to our proposed method IFL-GCL, and GRACE and GCA \cite{GRACE, GCA} are directly related to our proposed method IFL-GCL as they both utilize the InfoNCE loss. The directly-related baselines are used as warm-up. We denote our method based on GRACE as IFL-GR and the one based on GCA as IFL-GC. For all the baselines, we carefully tuned the hyper-parameters to achieve the best performances.

\textbf{Evaluation and Metric.} We focus on the node classification as the downstream task to show the quality of GCL's pre-trained representations.  After pre-training via GCLs, we fine-tune a task-specific classifier with supervision. Specifically, we divide the dataset into train, valid, and test sets in a 1:1:8 ratio for supervised fine-tuning. Each time, we pre-train the graph model once and repeat the supervised-fine-tuning of linear node classifier 3 times, then we calculate the average and standard deviation of the node classification accuracy on test set.

\subsection{IFL-GCL on the Framework of Graph Pre-training}
As shown in table \ref{tab:Main Experiments}, we employ 7 commonly used graph contrastive learning methods as baselines, among which GRACE and GCA, which use InfoNCE as the loss function, were selected as directly relevant baselines. 
Additionally, the 6 datasets on the left side of the table are used for testing in IID scenarios, while the 3 datasets on the right side are used for testing in OOD scenarios. The main conclusions that can be drawn from this table are as follows.

\textbf{(1) Our methods outperform the baselines.}
Our methods IFL-GR and IFL-GC occupy the best performances across all datasets apart from Computers. 
And our models rank second-best on all but PubMed and Photo. 
It is worth noting that our method is on par with the optimal performance on Computers, as well as the suboptimal performance on PubMed and Photo.

\textbf{(2) Our methods consistently enhance the capability of directly related models.} As indicated in the last two rows of Table 1, the improvement results on all datasets are positive, with the highest increase up to 9.05\%. This demonstrates that after using IFL-GCL to correct sampling bias, the performance of GRACE and GCA is stably enhanced across all datasets.

\textbf{(3) Our method shows more significant improvement under OOD scenarios.} Under OOD scenarios, our methods occupy the top and second-best results across all datasets, with  greater \(\Delta_{GC}\) and \(\Delta_{GR}\) than IID scenarios. This may be due to the reason that our method bridges the gap between different distributions by treating non-augmented samples under different distributions as positive ones, enabling the model to capture transferable knowledge across different distributions. 
As a result, IFL-GCL demonstrates good OOD generalization capability.

\subsection{\textbf{IFL-GCL on the Framework of LLM as Graph Enhancers}}
\begin{table*}[htbp]
    \centering
    \caption{Node classification results with 4 pre-trained LLMs as feature enhancers across 2 datasets. We show the accuracy, standard deviation, best and second best on each column, IFL's improvement compared to baselines the same way in Table \ref{tab:Main Experiments}. }\renewcommand{\arraystretch}{1.2} 
    \setlength{\tabcolsep}{4.5pt} 
    
    \begin{tabular}[t]{ccc||cc||cc||cc}
    \hline
    \multirow{2}{*} & \multicolumn{2}{c||}{\large{\textbf{Llama3.2-1B}}} & \multicolumn{2}{c||}{
    \large{\textbf{Llama3.2-3B}}} & \multicolumn{2}{c||}{
    \large{\textbf{Qwen2.5-0.5B}}} & \multicolumn{2}{c}{
    \large{\textbf{QWen2.5-1.5B}}} \\ 
    
    \multicolumn{1}{c}{} & 
    \multicolumn{1}{c}{\large{\textbf{Cora}}} & 
    \multicolumn{1}{c||}{\large{\textbf{CiteSeer}}} & 
    \multicolumn{1}{c}{\large{\textbf{Cora}}} & 
    \multicolumn{1}{c||}{\large{\textbf{CiteSeer}}} & 
    \multicolumn{1}{c}{\large{\textbf{Cora}}} & 
    \multicolumn{1}{c||}{\large{\textbf{CiteSeer}}} & 
    \multicolumn{1}{c}{\large{\textbf{Cora}}} & 
    \multicolumn{1}{c}{\large{\textbf{CiteSeer}}}\\
    \midrule
    
    \makecell[c]{\large{\textbf{GRACE}}\\}
    &\makecell[c]{\(84.34_{(0.61)}\)}
    &\makecell[c]{\(75.48_{(0.95)}\)}
    &\makecell[c]{\(84.64_{(0.64)}\)}
    &\makecell[c]{\(76.75_{(0.30)}\)}
    &\makecell[c]{\(84.07_{(1.08)}\)}
    &\makecell[c]{\(75.83_{(0.27)}\)}
    &\makecell[c]{\(84.09_{(0.49)}\)}
    &\makecell[c]{\(75.52_{(0.51)}\)}
    \\

    \makecell[c]{\large{\textbf{GCA}}\\}
    &\makecell[c]{\(84.71_{(0.39)}\)}
    &\makecell[c]{\(\textit{\textbf{76.10}}_{(0.95)}\)}
    &\makecell[c]{\(84.93_{(0.69)}\)}
    &\makecell[c]{\(76.63_{(0.23)}\)}
    &\makecell[c]{\(84.08_{(1.35)}\)}
    &\makecell[c]{\(76.04_{(0.56)}\)}
    &\makecell[c]{\(\textit{\textbf{84.47}}_{(0.35)}\)}
    &\makecell[c]{\(76.31_{(0.76)}\)}\\

    \midrule
    \makecell[c]{\large{\textbf{IFL-GR}}\\}
    &\makecell[c]{\(\textit{\textbf{84.79}}_{(0.68)}\)}
    &\makecell[c]{\(76.02_{(0.74)}\)}
    &\makecell[c]{\(\textit{\textbf{85.03}}_{(0.23)}\)}
    &\makecell[c]{\(\textit{\textbf{77.65}}_{(0.40)}\)}
    &\makecell[c]{\(\textit{\textbf{84.12}}_{(0.99)}\)}
    &\makecell[c]{\(\textit{\textbf{76.29}}_{(0.51)}\)}
    &\makecell[c]{\(84.34_{(0.58)}\)}
    &\makecell[c]{\(\textit{\textbf{76.42}}_{(0.41)}\)}
    \\

    \makecell[c]{\large{\textbf{IFL-GC}}\\}
    &\makecell[c]{\(\underline{\textbf{85.38}}_{(0.37)}\)}
    &\makecell[c]{\(\underline{\textbf{76.42}}_{(0.82)}\)}
    &\makecell[c]{\(\underline{\textbf{85.43}}_{(0.99)}\)}
    &\makecell[c]{\(\underline{\textbf{77.72}}_{(0.46)}\)}
    &\makecell[c]{\(\underline{\textbf{85.38}}_{(0.68)}\)}
    &\makecell[c]{\(\underline{\textbf{77.13}}_{(0.34)}\)}
    &\makecell[c]{\(\underline{\textbf{85.37}}_{(0.83)}\)}
    &\makecell[c]{\(\underline{\textbf{77.63}}_{(0.19)}\)}
    \\

    \midrule
    \makecell[c]{\large{\(\mathbf{\Delta_{GR}}\)}\\}
    &\makecell[c]{+0.45\%}
    &\makecell[c]{+0.54\%}
    &\makecell[c]{+0.39\%}
    &\makecell[c]{+0.90\%}
    &\makecell[c]{+0.05\%}
    &\makecell[c]{+0.46\%}
    &\makecell[c]{+0.25\%}
    &\makecell[c]{+0.90\%}
    \\

    \makecell[c]{\large{\(\mathbf{\Delta_{GC}}\)}\\}
    &\makecell[c]{+0.67\%}
    &\makecell[c]{+0.32\%}
    &\makecell[c]{+0.5\%}
    &\makecell[c]{+1.09\%}
    &\makecell[c]{+1.30\%}
    &\makecell[c]{+1.09\%}
    &\makecell[c]{+0.90\%}
    &\makecell[c]{+1.32\%}
    \\
    
    \bottomrule
    \end{tabular}
    
    \label{tab:llm_emb}
\end{table*}
LLM plays a significant role in the study of graph foundation models where using LLMs as feature enhancers for Text-Attributed-Graphs (TAGs) is a common paradigm. 
In this section, we select 2 TAGs Cora and CiteSeer and use 4 pre-trained LLMs Llama3.2-1B, Llama3.2-3B, Qwen2.5-0.5B and Qwen2.5-1.5B to encode the their nodes' raw text features into dense representations. 
Then we use the LLM-encoded-representations as input features for baselines GRACE and GCA, and our methods IFL-GR and IFL-GC, thereby testing the effectiveness of our method under the current trends in graph foundation model research. 
Based on the analysis of Table \ref{tab:llm_emb}, we can draw the following conclusions: 

\textbf{(1) Our method still outperforms directly-related baselines.} The positive values in both the \(\Delta_{GR}\) and \(\Delta_{GC}\) rows indicate that, our method consistently improves the performance of the directly related GCL baselines on downstream task across all four LLMs enhancers and two TAG datasets. Since the node features encoded by LLM can capture semantic information better than traditional shallow text features, the semantic-guided sampling approach proposed in this paper can more accurately identify similar non-augmented samples, resulting in superior performance.

\textbf{(2) Our method is sensitive to the choice of LLM and has the potential to conform to scaling laws.} Under the same other conditions, different LLMs yield varying results when used as enhancers. Specifically, Llama3.2-3B significantly outperforms Llama3.2-1B in all IFL-GR and IFL-GC results, with advantages of 1.3\% and 1.6\% on the CiteSeer dataset, respectively; Qwen2.5-1.5B is slightly better than Qwen2.5-0.5B. This phenomenon is due to our method's reliance on the modeling capability of representations for semantics; the larger the LLM, the stronger the enhanced features' ability to model semantics, which in turn facilitates our method's ability to find the correct \(D_U^+\). Therefore, based on theoretical analysis and preliminary experimental results, our method has the potential to exhibit stronger bias correction capabilities as the scale of the LLM used as an enhancer increases.

Overall, our semantic-guided sampling approach for GCL holds great potential for the current research trend where LLMs are commonly involved in constructing graph foundation models.

\subsection{Analysis of the Discovered Positive Samples and hyper-parameters}
We also conduct extensive empirical analysis and evaluation on the proposed semantic-guided sampling approach, including: the analysis of discovered semantically similar non-augmented samples, and the analysis of hyper-parameters.

\subsubsection{\textbf{Analysis of the Discovered Positive Samples}}
This section will analyze how the semantic similarity of the \(D_U^+\) samples identified by our method changes during training as it dynamically updates as shown in the following Fig.\ref{fig:UPos}. 
\begin{figure}[htbp]
    \centering
    \includegraphics[width=0.9\linewidth]{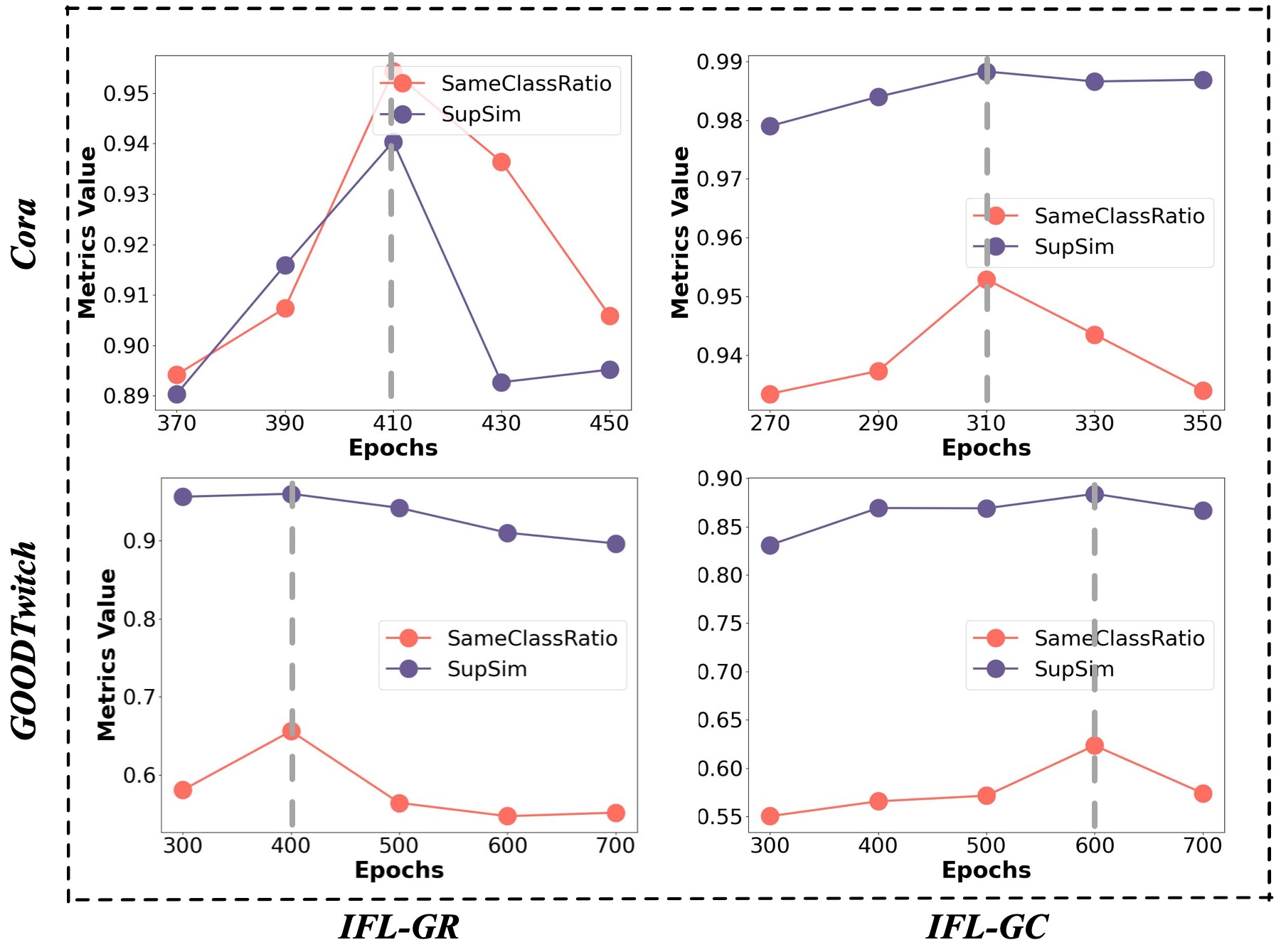}
    \caption{Analysis of semantic similarity of \(D_U^+\) during training. The gray dashed line indicates the training epoch of optimal checkpoint.}
    \label{fig:UPos}
\end{figure}
The x-axis of each subplot represents the number of completed training epochs including warming-up. And the interval on x-axis is set to the optimal update interval \(K\) for the corresponding setting. 
The y-axis shows the values of the two metrics measuring the average semantic similarity of node pairs in \(D_U^+\) found in the corresponding epoch. 
Besides, we mark the epoch of optimal checkpoint with gray dashed line to research the relationship between model performance and \(D_U^+\) semantic similarity during training process.
To consider both IID and OOD scenarios, we selected the Cora and GOODTwitch as datasets. We statistically measure the semantic similarity of the \(D_U^+\) samples found at each stage of the training process of IFL-GR and IFL-GC, respectively. 
We propose two metrics to reflect the similarity of elements in comparative samples: the \(SameClassRatio\) which is the proportion of samples with the same class label of \(D_U^+\), and the \(SupSim\) which is the average similarity of supervised representations of \(D_U^+\). 

Specifically, the \(SameClassRatio\) refers to: 
\begin{equation}
   SameClassRatio =  \frac{|\{(n,n')\}_{(n,n')\in D_U^+}^{Y_n = Y_{n'}}|}{|\{(n,n')\}_{(n,n')\in D_U^+}|}
\end{equation}
where \(Y_n\) means the class label of node \(n\) in node classification task, \(|\cdot|\) computes the number of contrastive samples in the set.

The supervised-representation-similarity \(SupSim\) use the graph encoder \(f_{\theta^*}\) trained with supervision as shown in section \ref{preliminary} to calculate the average representation similarity of all node pairs in \(D_U^+\) as shown in the following equation:
\begin{equation}
    SupSim = \frac{1}{|\{(n,n')\}_{(n,n')\in D_U^+}|} \sum_{(n,n') \in D_U^+} \frac{\mathbf{H}_n \cdot \mathbf{H}_{n'} }{\|\mathbf{H}_n\| \|\mathbf{H}_{n'}\|} 
\end{equation}
where \(\mathbf{H}\) is the nodes representation of \(\mathbf{G}\) encoded by \(f_{\theta^*}\),  \(\mathbf{H}_n\) is the representation of node \(n\),
\(\|\cdot\|\) is \(L_2\) norm of the vector. 
These metrics reflect the semantic proximity of the mined \(D_U^+\) samples from different perspectives; a higher metric indicates a higher probability of \(D_U^+\) being positive samples. 

Observing Fig.\ref{fig:UPos}, it can be seen that the trends of \(SameClassRatio\) and \(SupSim\) are largely consistent, both increase first and then decrease. And the peak value points of two metrics both correspond to the optimal checkpoint epoch.
This indicates that in the early stages, as the model trains to better and better state, the ability of GCL model gradually reaches optimality, leading to an overall increase in the semantic similarity of \(D_U^+\). In the meanwhile, better \(D_U^+\) samples correct the sampling bias in GCL.
As model training progresses, overfitting and representation collapse issues may cause the decline in the model's capability, resulting in the decrease of the \(D_U^+\) quality. Thus the semantic similarity of \(D_U^+\) declines after the best epoch.
This trend reflects a close relationship between the model's ability and the quality of \(D_U^+\), with the two aspects complementing each other. 
This indicates that IFL-GCL is capable of accomplishing sampling bias correction based on a warmed-up GCL model and enhancing it’s capabilities.

\subsubsection{\textbf{Analysis of Hyper-Parameters}}
\begin{figure}[htbp]
    \centering
    \includegraphics[width=\linewidth]{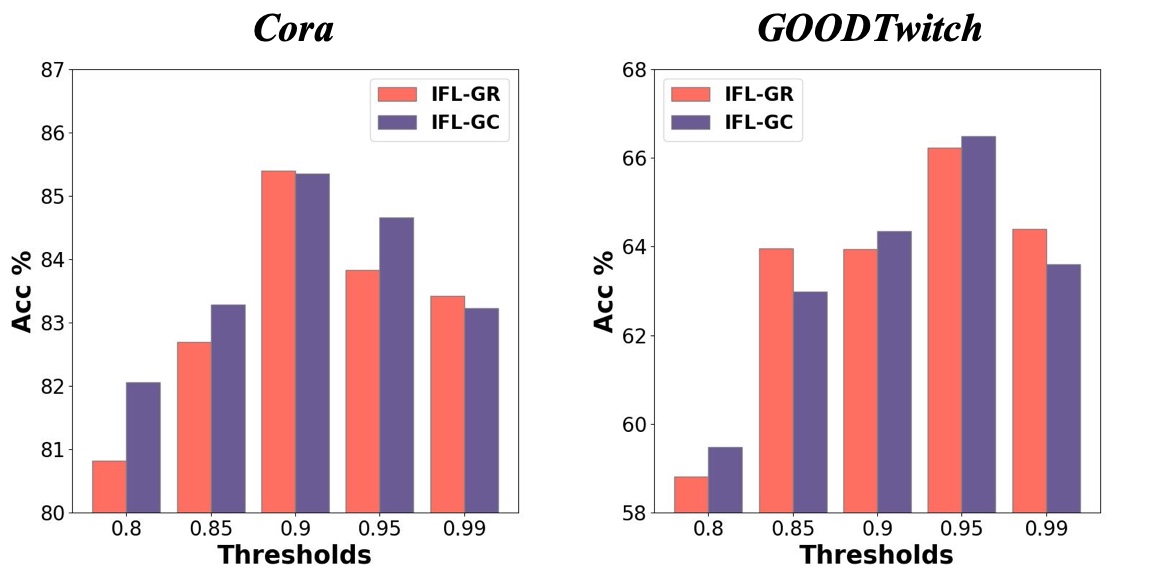}
    \caption{Analysis of Hyper-parameter: threshold \(t_s\)}
    \label{fig:threshold}
\end{figure}

In this part we study the effect of the following important hyper-parameters: warm-up epochs \(M\), update-interval epochs \(K\) and the threshold \(t_s\). 
\begin{figure}[htbp]
    \centering
    \includegraphics[width=\linewidth]{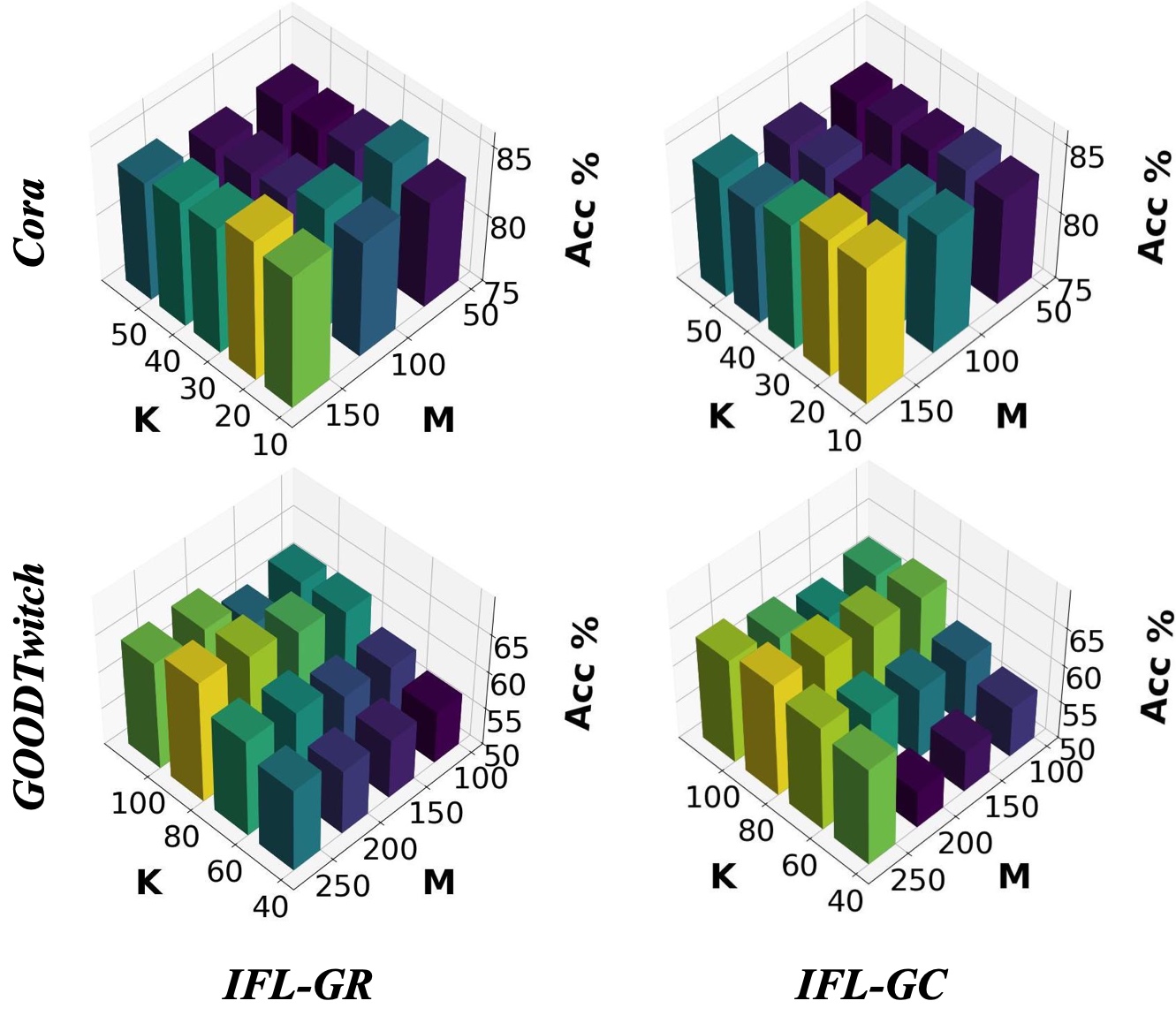}
    \caption{Analysis of Hyper-parameters: number of warm-up epochs \(M\) and number of update-interval epochs \(K\).}
    \label{fig:update-warmup}
\end{figure}
\begin{itemize}
    \item \textbf{Effect of warm-up and update-interval epochs.} As shown Fig.\ref{fig:update-warmup}, we conducted pre-training with different values of warm-up epochs \(M\) and update-interval epochs \(K\) on the Cora and GOODTwitch datasets using IFL-GR and IFL-GC, followed by the same supervised fine-tuning process on the same training set and report the node classification accuracy for each set. 
    Observing this 3D bar chart, it can be seen that for both datasets, the performance generally improves as \(M\) increases within a certain range. We believe this is because within a certain range, a larger \(M\) allows the learned \(s_{\theta}(\cdot, \cdot)\) to model \(r(\mathbf{x})\) more adequately, thereby capturing semantic information more effectively and performing better bias correction. Additionally, it can be observed that the update interval \(K\) varies for different datasets. 
    We think that it is related to the differences in the number of \(D_U^+\) across different datasets. GOODTwitch have far more nodes in total than Cora, leading to more nodes in \(D_U^+\) thus it needs larger update interval \(K\) to learn from these \(D_U^+\) samples for correcting the sampling bias.
    
    \item \textbf{Effect of threshold.} To study the impact of the threshold \(t_s\), we still conduct experiments on the Cora and GOODTwitch datasets for both IFL-GR and IFL-GC.
    With other hyper-parameters optimized, we set the threshold range to \{0.8, 0.85, 0.90, 0.95, 0.99\}, and the results are shown in Fig.\ref{fig:threshold}. 
    The experimental results indicate that our method is sensitive to the threshold \(t_s\), and different datasets have different optimal thresholds. 
    For Cora the optimal threshold is 0.9 and for GOODTwitch is 0.95. 
    Besides, we find out that a too low or too high threshold hurts the model's performance on both datasets. 
    Here is our explanation: a threshold that is too low may lead to negative samples with low semantic similarity being incorrectly converted into positive samples which not only fails to correct the sampling bias but actually exacerbates it. 
    This causes the representations learned by GCL can not accurately capture semantic information, leading to poor performance in downstream tasks.
    While a threshold that is too high may result in a too small number of \(D_U^+\) samples, which play a minimal role in correcting thus the model cannot be optimized to its best state. Therefore, both too high and too low thresholds can affect the bias correction performance of our method.
\end{itemize}

\section{Conclusion} 
This paper addresses the sampling bias issue in traditional graph contrastive learning by treating it as a Positive-Unlabeled learning problem where the definition of self-supervised tasks and contrastive samples should be semantically guided. We propose IFL-GCL, utilizing InfoNCE as a "free-lunch" to extract semantic information for resampling and redefine the maximum likelihood objective based on the corrected samples, resulting in a new InfoNCE loss function. Extensive experiments demonstrate the effectiveness and potential of our method in graph pretraining and graph foundation model research.

\section{Acknowledgment}
This work was supported by the Strategic Priority Research Program of the CAS under Grants No. XDB0680302 and the National Natural Science Foundation of China (Grant No.U21B2046, No.62202448).

\newpage
\bibliographystyle{ACM-Reference-Format}
\balance

\end{document}